\documentclass[10pt,final,twocolumn]{IEEEtran}

\usepackage{textcomp}
\DeclareSymbolFont{tildelow}{TS1}{cmr}{m}{n}
\DeclareMathSymbol{\tildelow}{0}{tildelow}{126}

\usepackage{bbm}
\usepackage{xcolor}
\usepackage{color}
\usepackage{graphicx}
\usepackage{epsfig}
\usepackage{subfigure}
\usepackage{amssymb}
\usepackage{amsbsy}
\usepackage{cite}
\usepackage{caption}
\usepackage{amsthm}
\usepackage{romannum}
\usepackage[noend]{algorithmic}
\usepackage{algorithm}

\newtheorem{theorem}{Theorem}
\newtheorem{lemma}{Lemma}

\newtheorem{definition}{Definition}

\newcommand{\beq}{\begin{equation}}
\newcommand{\eeq}{\end{equation}}
\newcommand{\bea}{\begin{array}}
\newcommand{\ena}{\end{array}}
\newcommand{\bds}{\begin {itemize}}
\newcommand{\eds}{\end {itemize}}
\newcommand{\bdf}{\begin{definition}}
\newcommand{\blm}{\begin{lemma}}
\newcommand{\edf}{\end{definition}}
\newcommand{\elm}{\end{lemma}}
\newcommand{\bthm}{\begin{theorem}}
\newcommand{\ethm}{\end{theorem}}
\newcommand{\bprp}{\begin{prop}}
\newcommand{\eprp}{\end{prop}}
\newcommand{\bcl}{\begin{claim}}
\newcommand{\ecl}{\end{claim}}
\newcommand{\bcr}{\begin{coro}}
\newcommand{\ecr}{\end{coro}}
\newcommand{\bquest}{\begin{question}}
\newcommand{\equest}{\end{question}}

\newcommand{\larrow}{{\larrow}}

\def\urltilda{\kern -.15em\lower .7ex\hbox{\~{}}\kern .04em}

\usepackage{booktabs}
\usepackage{tabularx}
\usepackage{hyperref}
\usepackage[margin=0.95in]{geometry}

\usepackage{float}
\usepackage[T1]{fontenc}
\usepackage{url}
\usepackage{ifthen}
\usepackage{cite}
\usepackage[cmex10]{amsmath} 
\usepackage{tikz}

\renewcommand{\baselinestretch}{1.06}

\begin{document}

\title{A Masked Pruning Approach for Dimensionality Reduction in Communication-Efficient Federated Learning Systems
}
\author{
    {Tamir L.S. Gez, Kobi Cohen (\emph{Senior Member, IEEE})}
    \thanks{T. Gez and K. Cohen are with the School of Electrical and Computer Engineering, Ben-Gurion University of the Negev, Beer-Sheva, Israel (e-mail:liorge@post.bgu.ac.il; yakovsec@bgu.ac.il).} 
	\thanks{This work was supported by the Israel Science Foundation under Grant 2640/20.}
 \thanks{Open-source code for the algorithm and simulations developed in this paper can be found on GitHub at \cite{gez2023MPFL_code}.}
 \thanks{This work has been submitted to the IEEE for possible publication. Copyright may be transferred without notice, after which this version may no longer be accessible.}
	\vspace{-0.75cm}
}

\maketitle
\pagenumbering{arabic}

\begin{abstract}

Federated Learning (FL) represents a growing machine learning (ML) paradigm designed for training models across numerous nodes that retain local datasets, all without directly exchanging the underlying private data with the parameter server (PS). Its increasing popularity is attributed to notable advantages in terms of training deep neural network (DNN) models under privacy aspects and efficient utilization of communication resources. Unfortunately, DNNs suffer from high computational and communication costs, as well as memory consumption in intricate tasks. These factors restrict the applicability of FL algorithms in communication-constrained systems with limited hardware resources.

In this paper, we develop a novel algorithm that overcomes these limitations by synergistically combining a pruning-based method with the FL process, resulting in low-dimensional representations of the model with minimal communication cost, dubbed Masked Pruning over FL (MPFL). The algorithm operates by initially distributing weights to the nodes through the PS. Subsequently, each node locally trains its model and computes pruning masks. These low-dimensional masks are then transmitted back to the PS, which generates a consensus pruning mask, broadcasted back to the nodes. This iterative process enhances the robustness and stability of the masked pruning model. The generated mask is used to train the FL model, achieving significant bandwidth savings. We present an extensive experimental study demonstrating the superior performance of MPFL compared to existing methods. Additionally, we have developed an open-source software package for the benefit of researchers and developers in related fields.
\end{abstract}

\section{Introduction}
\label{sec:intro}
In recent years, an increasing number of edge devices, including cell phones, IoT devices, autonomous vehicles, and others have integrated deep learning models into their functionalities. FL is an emerging ML paradigm designed for training models across multiple nodes, each holding local datasets, all without the need for direct exchange of sensitive data with a central PS. The rising popularity of FL can be attributed to its significant advantages in training DNN models while addressing privacy concerns and optimizing communication resource usage efficiently \cite{aledhari2020federated, kairouz2021advances, 9770266}. However, the drawback lies in the inherent challenges of DNNs, including high computational and communication costs, as well as memory consumption, particularly in complex learning tasks. These challenges pose limitations on the practical application of FL algorithms in communication-constrained systems with restricted hardware resources. 

\subsection{Dimensionality Reduction via Pruning Algorithms} 
\label{ssec:pruning_problem}

Various methods have been explored for low-dimensional representations of DNN models, including network quantization \cite{BinaryNet}, which minimizes memory storage requirements and expedites mathematical operations by converting weights into smaller units \cite{DoReFa_Net}. Similarly, the quantization of activation functions \cite{quantiztion_gaussian} reduces the resolution of specific layers, optimizing transition times. Another techniques, known as model distillation \cite{hinton2015distilling, livne2020pops}, involves transferring knowledge from a large, complex model (often referred to as the teacher model) to a smaller, simpler model (the student model). Recent studies have donstrated good performance of sparse convolutional model \cite{li2022revisiting}.  

In this paper, our emphasis lies on pruning-based methods, recognized as highly effective tools for diminishing the size of DNN models \cite{han2015learning, livne2020pops, chen2021only}. Pruning methods seek to reduce model size and operational complexity by eliminating connections or layers while maintaining performance. Pruning algorithms can be categorized based on the segment of the network they target for pruning, the criteria used to identify prune-worthy components, and the methods employed to execute pruning without compromising the network. These factors enable the identification of best pruning structures, criteria, and techniques. 

Determining which part of the network to prune involves two common approaches. Unstructured pruning, as introduced in \cite{han2016deep}, focuses on reducing the number of inter-layer connections to lower the floating-point operations per second required by the network. This method involves removing individual weights based on specific criteria, resulting in a sparser model. Conversely, structured pruning \cite{liu2017learning} targets more extensive structures such as entire neurons, layers, or channels. This approach preserves the original data layout and enhances hardware efficiency.

To identify parts eligible for pruning, a pruning criterion is established to assess the importance of individual parameters, filters, etc. This ranking may be determined by factors such as the magnitude of weights \cite{renda2020comparing} or the $L_1$ or $L_2$ norms \cite{zhuang2019discriminationaware} for structured pruning. Another option is to base the ranking on the magnitude of gradients \cite{blalock2020state}. Additionally, a decision must be made between pruning based on a global rank across the entire model or a local rank within each layer.

Lastly, to prune the network without compromising its performance, various common methods are employed. Some approaches involve training the full model, pruning based on the defined criterion, and subsequently fine-tuning the model with a smaller dataset \cite{han2015learning, liu2017learning}. Alternatively, some methods suggest retraining the model after pruning \cite{he2018soft, renda2020comparing}. The Lottery Ticket Hypothesis method \cite{frankle2019lottery} proposes pruning before training, eliminating the need for subsequent fine-tuning or retraining \cite{zhou2020deconstructing, morcos2019ticket}. Another technique, the Only Train Once method, introduces concepts such as zero-invariant groups and a unique optimization algorithm to achieve effective one-shot training and pruning \cite{chen2021only}. Additional strategies involve randomly creating a pruning mask over the model and adjusting this mask during training \cite{Mocanu_2018, mostafa2019parameter}. Some propose learning the pruning mask over the model when the cost function is based on the $L_0$ norm \cite{huang2018learning, he2019amc}. Penalty-based methods use the loss function to reduce weights during training \cite{Channel_Pruning, gao2019vacl}.

The studies mentioned above primarily concentrated on pruning algorithms within the framework of traditional centralized learning of DNN models. In these settings, the optimization for pruning is performed at a centralized unit with access to the entire dataset. However, the exploration of pruning methods within communication-constrained FL systems, where data is held locally at nodes without sharing, and each transmission of a generated model is resource-intensive, remains relatively unexplored. This paper specifically addresses this challenge.

\subsection{Communication-Efficient Federated Learning}
\label{ssec:fedlearn}

Federated learning can be characterized as a distributed learning problem within a communication network, encompassing numerous distributed nodes and a central PS. The primary goal of the PS is to address the following optimization problem:
\begin{equation}
\label{eq:weight_star_intro}
\boldsymbol{w}^* = \mbox{arg} \min_{\boldsymbol{w} \in W} \frac{1}{N}\sum_{n=1}^N f_n(\boldsymbol{w}),
\end{equation}
based on data signals received from the nodes. These data signals may result from local processing at the node, but do not involve direct access to the private and communication-intensive raw data. Here, $\boldsymbol{w}\in{W}\subset\mathbb{R}^d$ is the $d\times 1$ parameter vector which needs to be optimized. The solution $\boldsymbol{w}^*$ is known as the empirical risk minimizer. In ML tasks, we often write $f_n(\boldsymbol{w})=\ell(\boldsymbol{x}_n, y_n; \boldsymbol{w})$, which is the loss of the prediction on the input-output data pair sample $(\boldsymbol{x}_n, y_n)$, where $\boldsymbol{x}_n$ refers to the input vector and $y_n$ refers to the label, made with model parameter $\boldsymbol{w}$. 

Traditional ML algorithms solve (\ref{eq:weight_star_intro}) in a centralized manner. The conventional approach to ML entails storing all data in a central unit and utilizing a centralized optimization algorithm, such as gradient descent, to process the data. However, this method can be inefficient for data-intensive applications due to high bandwidth and storage requirements. FL offers a collaborative ML framework that addresses these issues by enabling distributed nodes to process and share a function of their locally-held data with a central PS, eliminating the need to upload the entire dataset \cite{konecny2015federated}. Importantly, this is achieved without the necessity for direct data exchange. This collaborative methodology addresses various challenges \cite{kairouz2021advances}, including the imperative to preserve privacy in AI applications and the growing demand to leverage distributed data sources. This approach is particularly well-suited for mobile applications, such as those in 5G, IoT, and cognitive radio systems, where communication resources are limited with digital  \cite{chen2020joint, abad2019hierarchical, naparstek2018deep, livne2020pops, shlezinger2020uveqfed, gafni2022distributed, gafni2022learning, ami2023client, salgia2023communicationefficient} or analog \cite{9076343, 9459539, 9562537, 9770266, gez2023subgradient} communications, and due to privacy concerns \cite{mcmahan2017communication, aledhari2020federated, 9042352}. Therefore, FL has emerged as a promising solution to the challenges faced by traditional centralized ML algorithms, garnering increasing attention in recent years.

The process of FL can be broken down into a few steps. First, the global model is shared with all distributed nodes. Each node then trains the model on its local data, creating a unique, local model update. The nodes communicate with a central PS to solve the optimization problem (\ref{eq:weight_star_intro}), sending their local output to the PS, which aggregates the data and updates the global model. This process is repeated over multiple rounds until the model's performance meets a predetermined criterion. FL offers a significant advantage in terms of user privacy, as the raw data remains confined to its original node, mitigating privacy concerns. Additionally, this distributed approach allows the utilization of extensive data generated across multiple nodes, potentially yielding highly robust and generalized models. Moreover, since most computations occur locally, data transmission is minimized, resulting in reduced bandwidth usage. In this paper, our focus is on developing a pruning algorithm over FL to generate low-dimensional representations of the models produced. This aims to provide efficient and powerful DNN models that can operate on resource-constrained edge devices with minimal communication costs.

\subsection{Main Results}
\label{ssec:contribution}

Exploring the development of pruning algorithms within a federated learning architecture introduces a new research domain. Recent studies focused on performing the pruning process on the central server or strong clients, neglecting to leverage the inherent structure of the FL problem. This approach incurs significant bandwidth requirements, as observed in recent works such as \cite{jiang2022model} and subsequent studies. This paper takes a progressive step in research by concentrating on communication-constrained FL systems. In these systems, the imperative is to complete the entire FL task with minimal communication consumption. 

To address this challenge, we develop a novel algorithm named Masked Pruning over Federated Learning (MPFL), designed to capitalize on the non-linearity of the pruning process in FL systems. First, instead of transmitting full precision weights, which may require 32 or 64 bits, MPFL communicates only a binary pruning mask. In the context of structured pruning, a single bit can succinctly represent an entire filter, significantly reducing the necessary channel bandwidth. Second, the approach demonstrates resilience to noise and contamination effects from outlier units. Through a collective voting mechanism on the joint pruning mask, MPFL achieves a high level of immunity, ensuring the integrity and performance of the shared model. In the MPFL framework, the pruning technique involves a non-linear operation across the channel. Each node independently learns from its data, and after several steps, a pruning mask is computed for each node based on the training process. Only the mask is shared, enabling a voting process and facilitating the exchange of the updated mask. This iterative process continues until the desired pruning level is attained.

In contrast to traditional pruning algorithms that depend on the global ranking of network weights using the entire dataset, as discussed in subsection \ref{ssec:pruning_problem}, our approach enables each unit to conduct local ranking and pruning. Many federated learning algorithms, as outlined in subsection \ref{ssec:fedlearn}, necessitate linearity in mathematical operations, a requirement our method does not impose. This distinction arises because our approach operates on a set of binary signals, where such linearity is not applicable. As detailed in Section \ref{sec:algo}, our approach shifts the focus from global information to localized data handling, thereby enhancing robustness against potential noise.

Second, to evaluate the proposed MPFL algorithm, we performed comprehensive simulation experiments utilizing VGG11 and ResNet18 models with CIFAR10 and ImageNet100 datasets, respectively. The outcomes validate the superior performance of MPFL compared to existing methods, particularly in the presence of noisy data or during high pruning steps. MPFL not only converges similarly to classical methods in standard scenarios but also surpasses them significantly under challenging conditions, achieving substantial bandwidth savings. These results highlight the robustness and effectiveness of the algorithm, signifying a substantial advancement in the field.

Third, we have created an open-source implementation of the MPFL algorithm, which is accessible on GitHub at \cite{gez2023MPFL_code}. Our experimental study showcases the versatility of the software across diverse challenging environments. We encourage the use of the open source software by researchers and developers in related fields.

\section{System Model and Problem Statement}
\label{sec:system}

We consider a federated learning system comprised of a parameter server (centralized unit) (PS) and a set of $N$ nodes (edge devices or clients). Each of the nodes (say node $n$) holds a local dataset of input-output samples $\{\boldsymbol{x}_{n}, y_{n}\}$ with $\{\boldsymbol{x}_{n}\}$ being the input vectors and $\{y_n\}$ the labels. The local datasets cannot be directly shared due to communication constraints or privacy considerations. The primary objective is to collaboratively train a deep learning model with dimension $d$ without exchanging the direct, raw data - given its costliness and privacy implications - while sharing only a condensed output from local processing. Furthermore, the dimension of the trained model is required to meet a dimension constraint such that $d\leq K$. Formally, the objective is to solve the following optimization problem:

\begin{equation}
\label{eq:main_problem_def}
\boldsymbol{w}^* = \mbox{arg} \min_{\boldsymbol{w} \in W \subseteq \mathbb{R}^d } \frac{1}{N}\sum_{n=1}^N f_n(\boldsymbol{w}) \quad \text{s. t.} \quad d\leq K.
\end{equation}
Here, $\boldsymbol{w}\in{W}\subseteq\mathbb{R}^d$ represents the $d\times 1$ parameter vector that needs to be optimized (often might represent a neural network model arranged in a vector), $K$ signifies the maximal allowed model dimension. The solution $\boldsymbol{w}^*$  is known as the empirical risk minimizer. The function $f_n(\boldsymbol{w})=\ell(\boldsymbol{x}_n, y_n; \boldsymbol{w})$ denotes the loss of prediction on the input-output local dataset computed with respect to the model parameter $\boldsymbol{w}$.

\noindent
\textbf{Bandwidth Efficiency:} 
As discussed in Section \ref{sec:intro}, numerous studies have proposed reducing dimensionality by pruning the model at the server based on received local models from the edge nodes. However, these methods often lack a specific focus on bandwidth efficiency, leading to a potentially high volume of transmitted data during the FL task between nodes and the PS. Conversely, a significant challenge in recent years within real-time FL systems is how to implement FL tasks with spectral-efficient considerations, aiming to reduce the total transmitted data in bits during the training procedure.

In this paper, we address this gap by specifically aiming to develop a spectral-efficient algorithm that solves \eqref{eq:main_problem_def}, reducing the total transmitted data in the network during the FL task. We evaluate the spectral efficiency performance of the algorithms by counting the total number of bits required to reach a specified accuracy. Our results demonstrate that our algorithm achieves state-of-the-art performance in this regard. 

\begin{figure}
    \centering
    \includegraphics[width=1.1\linewidth]{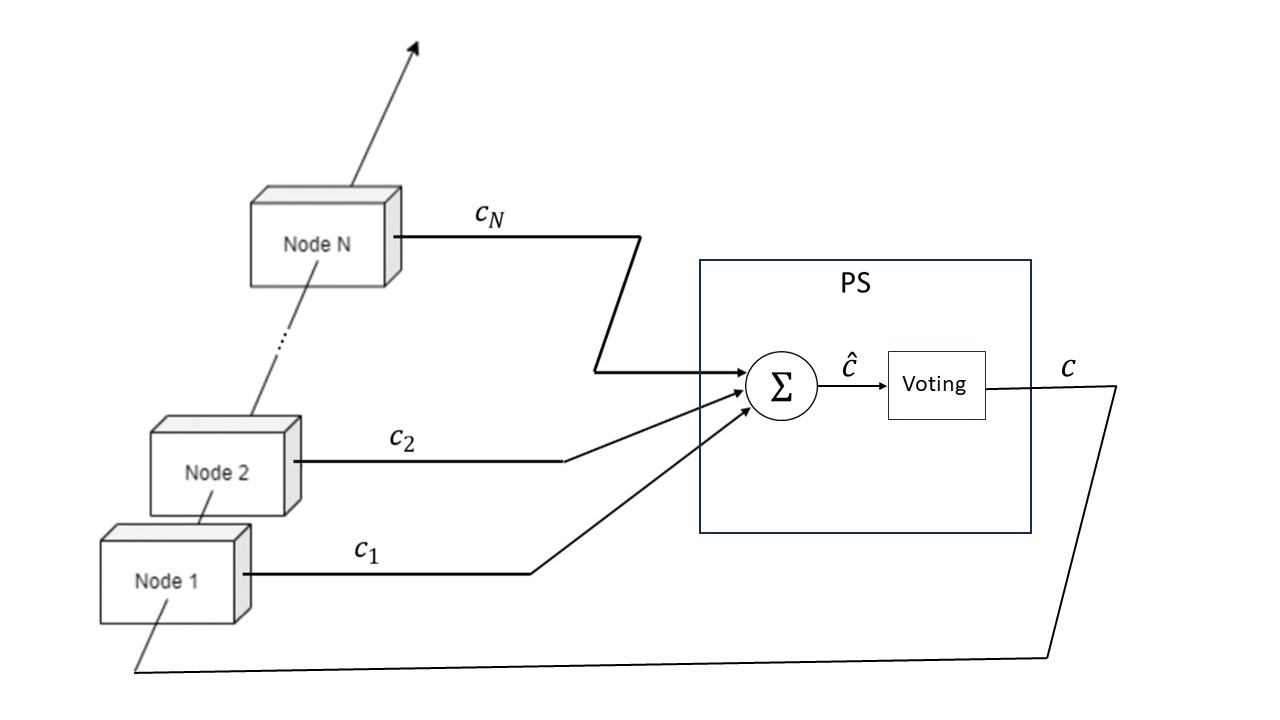}
    \caption{An illustration of the FL system.}
    \label{fig:system}
\end{figure}

\section{The Masked Pruning over Federated Learning (MPFL) Algorithm}
\label{sec:algo}

In this section, we present the MPFL algorithm used to solve the objective. To solve \eqref{eq:main_problem_def} in a federated manner while reducing the total amount of transmitted data used to accomplish the FL task, MPFL implements a novel distributed masked pruning approach. The key idea is to implement local pruning at each node locally, and then using masking technique to send non-zero locations of the trained local model. The mask signals are then aggregated at the server, which implements a voting technique to generate a global mask, which is then sent back to the nodes for the next iteration. This novel transmission design of masked-pruning signals enables the algorithm to generate FL model with low dimension while significantly saving the total amount of transmitted data. A description of the system is given in Figure \ref{fig:system}.

\subsection{Algorithm Description}

We will now elaborate on the detailed steps of the MPFL algorithm.\vspace{0.2cm} 

\noindent
\textbf{Initialization:} The PS initializes the algorithm by broadcasting an initial weight vector $\boldsymbol{w}_{0}$ to all nodes.\vspace{0.2cm}

\noindent
\textbf{Distributed Local Processing at the Nodes:} In the local processing step, each node solves locally the following optimization problem:
\begin{equation}
\label{eq:w_star}
\begin{array}{l}
\displaystyle
\boldsymbol{w}^*_{n} = \mbox{arg} \min_{\boldsymbol{w}_{n} \in W_{n}} \;\ell_n(\boldsymbol{x_{n}},{y_{n}}; \boldsymbol{w_{n}} \odot \boldsymbol{c_{n}})\vspace{0.2cm}\\\hspace{1.5cm}
s. t. \hspace{1cm}\|\boldsymbol{c}_{n}\|_{0}\leq K,
\end{array}
\end{equation}
where $\boldsymbol{w}_{n}$ is the local model weights of node $n$, $\boldsymbol{c}_{n}$ denotes the local mask of node $n$, containing entries with values one and zero, with $L_0$ norm smaller than $K$. The term $\ell_n$ denotes the local loss of node $n$ computed with respect to 
$\boldsymbol{w_{n}} \odot \boldsymbol{c_{n}}$, where $\odot$ refers to product element wise between the two vectors. 

We will demonstrate later that the transmitted data signals and federated aggregation operate exclusively on the mask signals for pruning the model, resulting in significant bandwidth savings. Each node trains its model locally using its unique datasets and the updated mask in \eqref{eq:w_star} to produce the updated model $\boldsymbol{w}{n}^{*}$. The process of generating the local pruning mask $\boldsymbol{c}_{n}$ is described next.

Let $M$ be the number of layers in the DNN, and $L(m)$ be the number of weights in layer $m$. Let $\boldsymbol{w}_n(m,l)$ denote the $\ell$th weight in the $m$th layer of the DNN model of node $n$. To compute the pruning mask, each node assigns score $\boldsymbol{s}_n(m,l)$ for each weight $\boldsymbol{w}_n(m,l)$ in the model. 
It is common to use $p$-norm with respect to the weights directly: 
\begin{equation}
\label{eq:weight_scroting}
s_n(m,l) =\|\boldsymbol{w}_n(m,l)\|_{p},
\end{equation}
or with respect to the gradients:
\begin{equation}
\label{eq:weight_grad_scroting}
s_n(m,l) =\left\|\frac{\partial{L}}{\partial{\boldsymbol{w}_n(m,l)}}\right\|_{p}.
\end{equation}

For example, consider a layer (say layer $m$ in \eqref{eq:weight_scroting}, \eqref{eq:weight_grad_scroting}) of a convolutional network containing a filter of size $64\times 3\times 3$, that is, $64$ filters of size $3\times 3$ (i.e., $1\leq l\leq 64$ in \eqref{eq:weight_scroting}, \eqref{eq:weight_grad_scroting}). Then, $s_{m,l}$ is a scalar score determined by the $p$-norm with respect to the $l$th weight matrix of size $3\times 3$. As a result, we obtain a score vector $\boldsymbol{s}_n(m)$ of size $64\times 1$ for the $m$th layer. In the experiments detailed in this paper, we utilized the scoring mechanism in \eqref{eq:weight_scroting} with $p=2$, i.e., $L_{2}$ norm, over each filter and layer.

The scores for all weights within the layer $m$ are then aggregated into a score vector $\boldsymbol{s}_n(m)$. This vector, of size $L(m) \times 1$, effectively represents the importance of each weight in the layer:
\begin{equation}
\boldsymbol{s}_n(m) = [s_n(m,1), s_n(m,2), \ldots, s_n(m,L(m))]^T.
\end{equation}

To form a comprehensive scoring mechanism for the entire DNN model, these layer-wise score vectors $\boldsymbol{s}_n(m)$ are concatenated. The resulting model-wide score vector $\boldsymbol{s}_n$ is a concatenation of all the layer-wise vectors, capturing the significance of each weight across the entire network:
\begin{equation}
\label{eq:vector_scoring}
\boldsymbol{s}_n = [\boldsymbol{s}_n(1)^T, \boldsymbol{s}_n(2)^T, \ldots, \boldsymbol{s}_n(M)^T]^T.
\end{equation}

Here, $\boldsymbol{s}_n$ is a vector with a length of $\sum_{m=1}^M L(m)$, where each part, $\boldsymbol{s}_n(m)$, provides scores for layer $m$. This comprehensive vector $\boldsymbol{s}_n$ aids in the pruning process by identifying those weights in the network that are least contribute to the model’s performance, thereby facilitating efficient model size reduction without significantly compromising accuracy.

Next, the algorithm in the local processing phase prunes the model according to the score of each layer (and not according to global scores). Each time the algorithm trims $x\%$ from each layer, the output of the mask $c_n(m,l)$ is a binary value $0$ for trimmed weight and $1$ for untrimmed weight. The pruning mask for each weight in layer $m$ of node $n$ is determined by:
\begin{equation}
\label{eq:mask_calculation}
\begin{array}{l}
c_n(m,l) =
\begin{cases}
    1,& \text{if } \|s_n(m,l)\|_{p}\geq th_n, \\
    0,& \text{otherwise},
\end{cases}
\end{array}
\end{equation}
and the mask values within the layer $m$ are then aggregated into a mask vector $\boldsymbol{c}_n(m)$. Here, $th_n$ represents the pruning threshold, computed according to the target sparsity level for the layer. For example, to prune $10\%$ of the weights in a layer, the sparsity factor is set to $0.1$. The threshold $th_n$ is subsequently determined by selecting the score that corresponds to the $10\%$ percentile in the distribution of scores $\boldsymbol{s}_n(m)$. This method ensures that only the least significant weights, as identified by their scores, are pruned to attain the desired sparsity level. 

As for the previous example, the score vector of size $1\times 64$ is now represented by a binary vector $\boldsymbol{c}_n(m)$ of size $64$ bits. In terms of data transmission volume, applying existing pruning algorithms in FL systems requires sending the layer of size $64\times 3\times 3$ to the server for global processing and pruning, which consumes  $b\times 64\times 3\times 3$ bits, where $b$ is the number of bits used to represent floating-point numbers (e.g., $b=32$ in float32 or $b=64$ in float64). As a result, the proposed MPFL algorithm requires only $1/9b$ transmission data at each iteration (less than $1\%$ compared to existing methods for typical float precision). As we will demonstrate later, the masked pruning approach implemented by MPFL almost does not harm the performance per iteration. Consequently, the overall performance throughout the entire execution of the algorithm remains robust, resulting in significant savings in data transmission for the same level of accuracy compared to existing methods.\vspace{0.2cm}, 

\noindent
\textbf{Uplink Transmissions to the PS:} 
At each iteration's uplink time, each node (say node $n$) transmits its local pruning masks $\boldsymbol{c}_{n}(m)$, $m=1, ..., M$, to the PS.\vspace{0.2cm} 

\noindent
\textbf{Global Processing at the PS:}
The PS receives the pruning masks from all nodes and compute an average pruning mask for each layer $m$ in the mask:
\begin{equation}
\label{eq:fed_mask}
\boldsymbol{\hat{c}}(m) = \frac{1}{N}\sum_{n=1}^{N}\boldsymbol{c}_{n}(m).
\end{equation}

Next, the algorithm creates a consensus histogram across the edge devices. As the pruning level is increased, the consensus among the devices generally diminishes. Therefore, given the generated histogram, we adopt two strategies: The first requires a minimum pruning level for each layer, and the second involves a dynamic threshold based on the histogram. The PS selects the global mask of layer $m$, $\boldsymbol{c}(m)$ such that $\|\boldsymbol{c}(m)\|_{0}<=K$ out of $\boldsymbol{\hat{c}}(m)$, where $L_{0}$ is the zero norm, i.e. by counting how many elements are different than zero, and $K$ is the amount of element allowed keeping at most per layer. For example, if $100,000$ parameters are given in the model, and it is desired to prune $10\%$ of the DNN, then we require $K$ to be $90,000$. Finally, the PS obtains a pruning mask $\boldsymbol{c}$ concatenating the masks of all layers, $\boldsymbol{c}(m)$, $m=1, ..., M$. \vspace{0.2cm}

\noindent
\textbf{Downlink Transmissions to the Nodes:}
Upon the generation of a collaborative mask $\boldsymbol{c}$, the PS broadcasts $\boldsymbol{c}$ to all nodes for the next iteration. Once the maximal pruning level has been achieved, the direct low-dimensional model is then transmitted and aggregated by standard FL iterations for global aggregation of all nodes.

It is worth noting that as per initial studies referenced in Section \ref{ssec:pruning_problem}, it is more effective to prune incrementally rather than all at once. For example, pruning 50\% of the network is better accomplished by five rounds of 10\% pruning rather than a single 50\% pruning step. 

\begin{figure}
    \centering
    \includegraphics[width=0.9\linewidth]{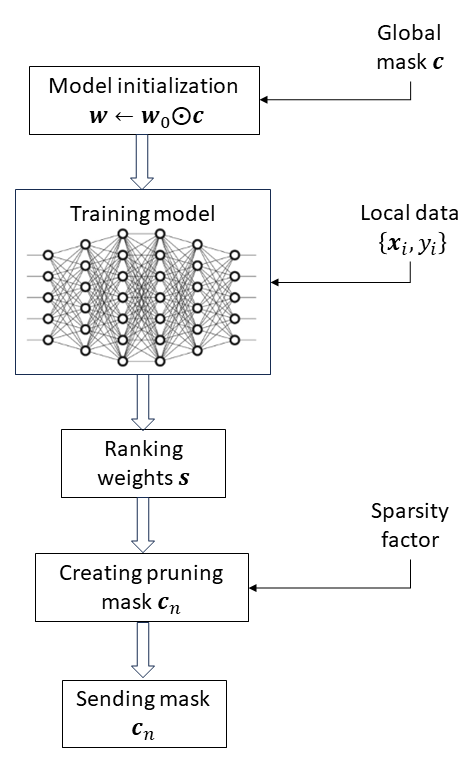}
    \caption{An illustration of the local processing steps at each node in the MPFL algorithm.}
    \label{fig:edge_algorithm}
\end{figure}

\begin{figure}
    \centering
    \includegraphics[width=0.7\linewidth]{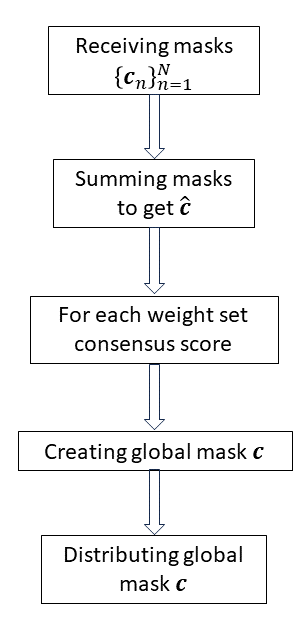}
    \caption{An illustration of the global processing steps at the PS in the MPFL algorithm.}
    \label{fig:ps_algorithm}
\end{figure}

The pseudocode for the MPFL algorithm is provided in Algorithm \ref{alg:algo}. Block-diagram illustrations for the operations at the nodes and the PS are provided in Figs. \ref{fig:edge_algorithm}, \ref{fig:ps_algorithm}, respectively.

\begin{algorithm}
\footnotesize
\caption{The MPFL Algorithm}\label{alg:algo}
\begin{algorithmic}[1]
    \STATE \textbf{initializing:} The PS broadcasts $\boldsymbol{w}_{0}$ for all nodes
    \FOR {iteration $k=0, 1, ...$}
        \STATE Each node trains a local model to get local optimized weights $\boldsymbol{w}_{n}^{*}$
        \STATE Each node computes the local pruning scores $\boldsymbol{s}_n$ using \eqref{eq:vector_scoring}
        \STATE Each node computes the local pruning mask $\boldsymbol{c}_{n}$ using (\ref{eq:mask_calculation})
        \STATE Each node transmits its local pruning mask $\boldsymbol{c}_{n}$ to the PS
        \STATE PS receives all local pruning masks ${\{\boldsymbol{c}_{n}\}}_{n=0}^N$
        \STATE PS computes an average pruning mask per layer $\hat{\boldsymbol{c}}(m)$ using \eqref{eq:fed_mask}
        \STATE PS computes a consensus histogram per layer $\boldsymbol{c}(m)$ using $\hat{\boldsymbol{c}}(m)$
        \STATE PS updates the global pruning mask $\boldsymbol{c}$ based on the histogram
        \STATE PS broadcasts $\boldsymbol{c}$ to the nodes
    \ENDFOR (Repeat the process for incremental pruning sparsity and enhanced network stability)
\end{algorithmic}
\end{algorithm}

\subsection{Further Insights and Observations of MPFL Algorithm}

\noindent
\textbf{Non-Linearity and Error Mitigation:} Consider an FL system with $N$ nodes, where each node has a set of weights denoted by $\boldsymbol{w}_{n}$. As per the federation model described in subsection \ref{ssec:fedlearn}, the aggregated model weights can be expressed by:
\begin{equation}
\label{eq:fl}
\boldsymbol{w} = \frac{1}{N}\sum_{n=1}^{N} \boldsymbol{w}_{n}.
\end{equation}

Next, we define the score assigned to each weight in (\ref{eq:scors}) below based on the definition in (\ref{eq:vector_scoring}, In this example, we use magnitude scoring, but the same argument holds for gradient scoring.
\begin{equation}
\label{eq:scors}
\begin{array}{l}
\boldsymbol{s} = \|\boldsymbol{w}\|_{p},  \vspace{0.2cm}\\
\boldsymbol{s}_{n} = \|\boldsymbol{w}_{n}\|_{p}.
\end{array}
\end{equation}

Applying the triangle inequality yields:
\begin{equation}
\label{eq:nonelinscore}
\begin{array}{l}
\displaystyle\boldsymbol{s} = \|\boldsymbol{w}\|_{p} = \|\frac{1}{N}\sum_{n=1}^{N}\boldsymbol{w}_{n}\|_{p} \vspace{0.2cm}\\ 
\displaystyle
\leq\frac{1}{N}\sum_{n=1}^{N}\|\boldsymbol{w}_{n}\|_{p} = \frac{1}{N}\sum_{n=1}^{N}\boldsymbol{s}_{n}.
\end{array}
\end{equation}

Thus, when defining a pruning mask as in (\ref{eq:mask_calculation}), we observe a lack of linearity:
\begin{equation}
\label{eq:nonlineq}
\begin{array}{l}
\displaystyle\boldsymbol{c(m,l)} =
\begin{cases}
    1,& \text{if } s(m,l) \geq th \\
    0,& \text{else}
\end{cases} \\ 
\displaystyle=
\begin{cases}
    1,& \text{if } \|\boldsymbol{w}(m,l)\|_{p} \geq th \\
    0,& \text{else}
\end{cases}  \\ 
\displaystyle=
\begin{cases}
    1,& \text{if } \|\sum_{n=1}^{N}{\boldsymbol{w}(m,l)}\|_{p} \geq th \\
    0,& \text{else}
\end{cases}
 \\ 
 \displaystyle\neq
\frac{1}{N}\sum_{n=1}^{N}
\begin{cases}
    1,& \text{if } \|\boldsymbol{w}_{n}(m,l)\|_{p} \geq th \\
    0,& \text{else}
\end{cases} \\ 
\displaystyle=
\frac{1}{N}\sum_{n=1}^{N}
\begin{cases}
    1,& \text{if } s_n(m,l)\geq {th}_{n} \\
    0,& \text{else}
\end{cases}
\\ 
\displaystyle= \frac{1}{N}\sum_{n=1}^{N}c_{n}(m,l) = \hat{c}(m,l),
\end{array}
\end{equation}
where $\hat{\boldsymbol{c}}$ is the average mask as described in (\ref{eq:fed_mask}), and $th$ is the threshold computed based on the required sparsity factor as described in (\ref{eq:mask_calculation}).

In scenarios where a node generates noise or errors, a typical federated model would converge in a manner similar to a centralized model. This convergence could lead to the inadvertent pruning of crucial weights. However, in MPFL, each node contributes equally to the voting process with a weight of $1/N$. Consequently, a single contaminated node will be identified and filtered out as an outlier. This feature demonstrates the robustness of MPFL in handling noise and disturbances.\vspace{0.2cm}

\noindent
\textbf{Bandwidth Efficiency:} We will now illustrate the communication savings per iteration achieved by the MPFL algorithm, which leverages the transmission of masked pruning instead of the pruned weights directly. To elaborate, if we consider a model employing unstructured pruning, we typically allocate $b$ bits for each weight in the network (e.g., float 32 or float64). In this scenario, MPFL achieves communication savings equivalent to $b$ times the required communication volume. Second, in practical DNNs employing models with structured pruning, MPFL assigns a single bit to each filter. For example, in a convolutional layer of size $L\times R\times Q$, where each filter kernel is of size $R\times Q$ and there are $L$ filters, MPFL utilizes only $L$ bits for mask transmissions for this layer. This is in contrast to the conventional approach that would use $b\times L\times R\times Q$ bits (where $b$ is the number of bits used to represent floating-point numbers). Consequently, MPFL achieves significant savings by a factor of $b\times R\times Q$ bits.

For example, consider a VGG16 network, which consists of $16$ layers. The architecture is as follows: The first and second layers are convolutional filter type of size $64\times 3\times 3$. The third and fourth layers are convolutional filter type of size $126\times 3\times 3$. The next three layers are convolutional filter type of size $256\times 3\times 3$. The next six layers are convolutional filter type of size $512\times 3\times 3$. The last three layers are fully connected type of size $4096$. Therefore, in traditional FL algorithms with float64, at each uplink transmission time every node is required to transmit: 
$2*3*3*64*64 + 2*3*3*128*64 + 3*3*256*64 + 6*3*3*512 + 3*4,096*64 = 1,182,720$ bits. By contrast, in MPFL, every node is required to transmit: $2*64+2*128+3*256+6*512+3*4096 = 16,512$ bits only. As a result, MPFL achieves a savings of approximately $98.6\%$ in communication volume per iteration while maintaining strong performance, as illustrated in the simulation results in Section \ref{sec:performance}.

\section{Experiments}
\label{sec:performance}

In this section, we present a series of simulations conducted to evaluate the performance of our proposed algorithm. We simulated an FL system with $N$ nodes and a PS. The dataset was distributed across the nodes in each experiment to simulate the scenario of distributed nodes storing local data. The central unit performs computations at the PS using MPFL. We compared the following algorithms: (i) \emph{pruning-FL:} The pruning-FL algorithm was recently proposed to generate a low-dimensional pruned model in FL systems and is commonly used for this purpose today \cite{jiang2022model}. In pruning-FL, the model is learned in a federated manner, and the pruning update is implemented at the PS during the algorithm; (ii) \emph{LTH pruning algorithm:} The Lottery Ticket Hypothesis (LTH) algorithm creates a pruning mask that would have been obtained from centralized training with direct access to the entire dataset in the context of classical pruning methods \cite{savarese2021winning}; and (iii) the proposed MPFL algorithm. 

When implementing pruning-FL and MPFL, the communication consumption encompasses the total data transmitted between the nodes and the PS during all iterations of the algorithms. In the case of LTH, we calculate the communication consumption as a one-shot uploading of the entire dataset (which is much more costly than the low-dimensional representations transmitted at each iteration in a federated manner) from the nodes to the PS. From this point onward, the computation is performed at the PS, with no additional communication consumption between nodes and the PS.

Our findings indicate that in a system with identical and independent distributed data and no noise or interference, the three algorithms converged similarly, with the proposed MPFL algorithm showing a slight advantage when the size of the network pruning was large, helping prevent layer resets (a side effect of network pruning), but with significant bandwidth savings by MPFL. It should be noted that MPFL performed well when presented with a noisy node that contaminated the environment by providing noisy samples or mislabels, maintaining robustness with respect to contamination. Additionally, the amount of information sent over the channel, i.e., the required bandwidth, is almost zero compared to the other methods. The focus of this section is to present these main findings. Open-source code for the simulations can be found on GitHub at \cite{gez2023MPFL_code}.

\subsection{Experiment 1: Learning VGG11 Model}

In this simulation, we used the VGG11 \cite{simonyan2015deep} model with CIFAR10 dataset \cite{Krizhevsky2009LearningML}. We simulated $10$ distributed nodes and distributed the data among them. Subsequently, we artificially contaminated one of the nodes by adding noise to the information. Additionally, we randomly changed the labels in another node (for example, label 1 became 3, etc.).

\begin{figure}
    \centering
    \includegraphics[width=1.0\linewidth]{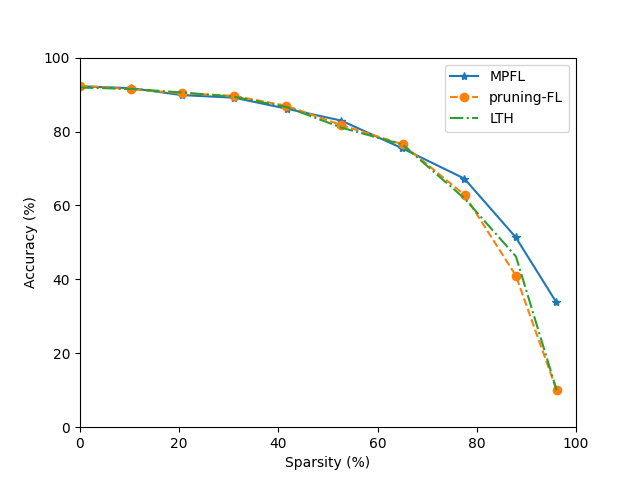}
    \caption{The accuracy versus sparsity levels obtained in Experiment 1 for pruning the VGG11 model using the CIFAR10 dataset.}
    \label{fig:vgg_prune}
\end{figure}

\noindent
\textbf{Accuracy Performance versus Sparsity Levels:} We start by presenting the accuracy performance versus the sparsity levels obtained by the algorithms. We executed the algorithms for network pruning until encountering a network performance reset, a common side effect resulting from excessive pruning in specific layers. Notably, our algorithm exhibited greater resilience to this phenomenon, accompanied by an improvement of a few percentage points when the network's pruning sparsity was high. An intriguing observation was that our network continued to yield results even after reducing over $90\%$ of its structure, an extent at which both LTH and pruning-FL algorithms reached the limit of the network's viability. These findings are illustrated in Figure \ref{fig:vgg_prune}.

\begin{figure}
    \centering
    \includegraphics[width=1.0\linewidth]{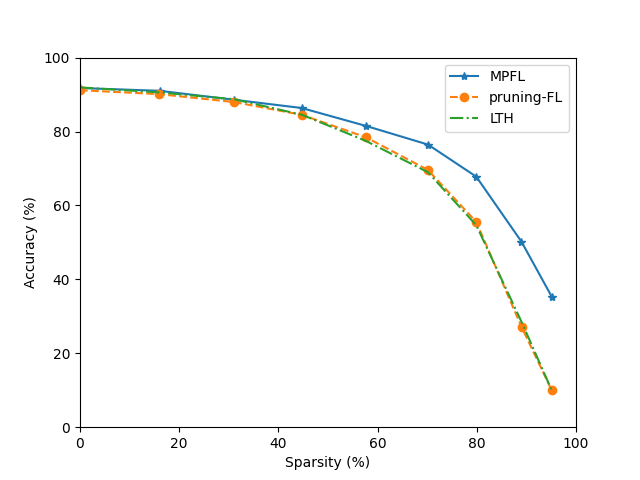}
    \caption{The accuracy versus sparsity levels obtained in Experiment 1 for pruning the VGG11 model using the CIFAR10 dataset. In these simulations, pruning is performed under noised input conditions and with shuffled labels.}
    \label{fig:vgg_prune_noised}
\end{figure}

Subsequently, we repeated the simulations, introducing noise to the input of one node. Additionally, for another node, we shuffled the tags, whereby tag 1 became tag 3, tag 2 became tag 1, and so forth. As anticipated, our algorithm demonstrated higher noise stability in this scenario compared to the other tested methods. These results are depicted in Figure \ref{fig:vgg_prune_noised}.

\begin{figure}
    \centering
    \includegraphics[width=1.0\linewidth]{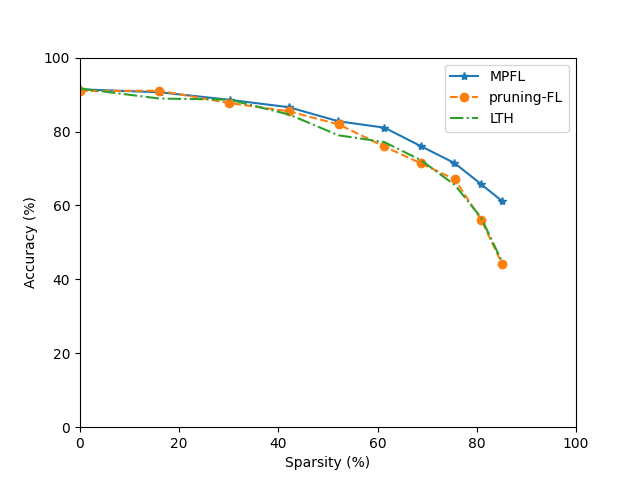}
    \caption{The accuracy versus sparsity levels obtained in Experiment 1 for pruning the VGG11 model using the CIFAR10 dataset. In these simulations, pruning is performed under noised input conditions and with shuffled labels using the histogram method.}
    \label{fig:vgg_hist_prune}
\end{figure}

Following, we replicated the simulations using the second method, which involves examining the histogram and transferring high histogram percentages, requiring a minimum of $90\%$ agreement within the histogram. This iteration yielded comparable results. Specifically, MPFL generated a mask with minimal impact on the network's performance, persisting in delivering results even after LTH and pruning-FL reached the limit of the network's viability. The results for these simulations are illustrated in Fig. \ref{fig:vgg_hist_prune}. It is noteworthy that we continue to achieve superior results, and with this method, we have additional parameters at our disposal to control errors across varying sparsity levels.

\noindent
\textbf{Bandwidth-Efficiency:} In Table \ref{tab:sim1_bandwidth}, we assess the data transmission volume among the algorithms. LTH operates centrally, involving a singular transfer where the entire communication-costly, high-dimensional raw data is uploaded to the parameter server (PS), followed by algorithm execution at the PS. We provide the total data transmission amount for this process. It is crucial to note that LTH does not address privacy concerns, requiring the non-sharing of direct data. It is presented solely as a performance benchmark. In contrast, MPFL and pruning-FL operate in a federated manner. They keep data locally at nodes while transmitting the updated model to the PS at each iteration. Here, we present the average transmitted data per iteration summed across all nodes, along with the overall data transmitted throughout all iterations in the entire simulation runs. 

Specifically, the centralized approach requires the transmission of all raw data, i.e., sending $60K$ images with a shape of $(32,32,3)$, equating to $17.6$ MB (megabytes). For pruning-FL, each transmission involves the model weights, totaling $128,807,306$ parameters for VGG11. Assuming the use of float32, this results in $491.36$ MB of transmitted data for each node per iteration. This calculation considers only the bytes required to send weights and biases without additional information for each node. In $10$ iterations of pruning, we send a maximum of $4.79$ GB for each node. By not resending pruned weights and with a total pruning rate not always at $10\%$ each iteration, the total bandwidth used amounts to $2.63$ GB for each node and $26.3$ GB overall.

In MPFL, each node sends only a bit for each filter and a bit per value in the bias layer, amounting to a total of $21,908$ bits per iteration from each node. This is equivalent to $2,739$ bytes for each iteration from each node, or $2.67$ KB from each node (where $KB$ refers to $1,024$ bytes in standard binary units). The bandwidth consumption required for the entire process is $27,390$ bytes for each node, i.e., less than $26.75$ KB for the entire process for each node. By saving bandwidth without resending weight masks for layers that have already been pruned, we achieve $15,064$ bytes (or $14.7$ KB) for each node and $147$ KB for the entire network. As can be seen, even for a relatively not too large dataset used in this experiment, MPFL offers significant bandwidth savings over the centralized approach, with better accuracy performance. The pruning-FL algorithm, in contrast, consumes significantly more bandwidth in this case.

\begin{table}[htbp]
    \centering
    \caption{Total amount of transmitted data in
Experiment 1 for pruning the VGG11 model using the
CIFAR10 datase}
    \label{tab:sim1_bandwidth}
    \begin{tabularx}{\columnwidth}{Xcc}
        \toprule
        Algorithm & Single iteration (max) & Full simulation \\
        \midrule
        LTH & - & 17.6 MB\\
        pruning-FL & 4.91 GB & 26.3 GB \\
        \textbf{Proposed MPFL} & \textbf{26.7 KB} & \textbf{147 KB} \\
        \bottomrule
    \end{tabularx}
\end{table}

\subsection{Experiment 2: Learning ResNet18 Model}

In these simulations, we used the ResNet18 model \cite{he2015deep} with ImageNet100 dataset, which is a subset of ImageNet dataset \cite{5206848}. This dataset is considerably larger, and consequently, we anticipate that a centralized approach like LTH, which uploads the raw data directly to the server, will consume significantly more bandwidth than MPFL and pruning-FL. The initial process was configured similarly to the previous simulation. In this simulation, our pruning strategy aimed to maximize dimensionality reductions without causing severe model degradation.

\begin{figure}
    \centering
    \includegraphics[width=1.0\linewidth]{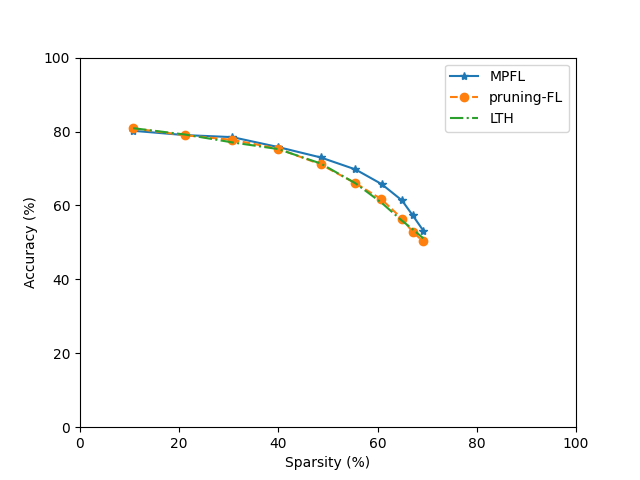}
    \caption{The accuracy versus sparsity levels obtained in Experiment 2 for pruning the ResNet18 model using the ImageNet100 dataset.}
    \label{fig:regular_compartion_10_units}
\end{figure}

\noindent
\textbf{Accuracy Performance versus Sparsity Levels:} We start by presenting the accuracy performance versus the sparsity levels obtained by the algorithms. Initially, we evaluated the three algorithms using 10 nodes, and the outcomes for accuracy versus sparsity levels are illustrated in Figure \ref{fig:regular_compartion_10_units}. Notably, MPFL demonstrates highly promising and successful results, outperforming the other methods. Particularly in higher pruning stages relative to the architecture, our algorithm maintains a superior accuracy compared to the alternatives.

\begin{figure}
    \centering
    \includegraphics[width=1.0\linewidth]{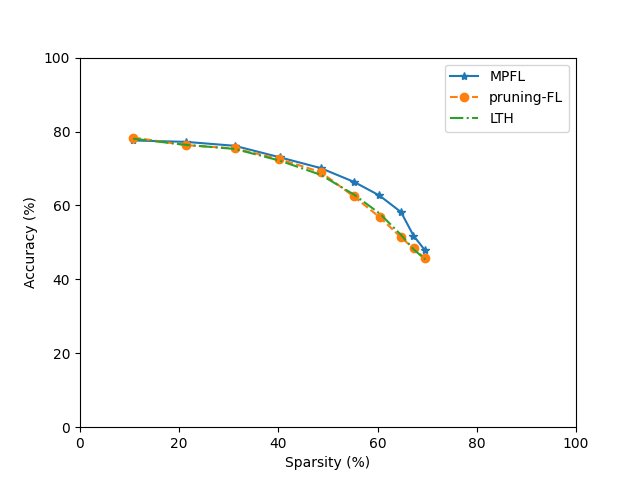}
    \caption{The accuracy versus sparsity levels obtained in Experiment 2 for pruning the ResNet18 model using the ImageNet100 dataset. In these simulations, pruning is performed under noised input conditions and with shuffled labels.}
    \label{fig:10_units_with_noise}
\end{figure}

Subsequently, we introduced noise by perturbing the input data of one node, and for another node, we shuffled the labels. Here again, we observe an advantage, although it is noteworthy that ResNet exhibits greater stability to such noises and disturbances. The change is not as pronounced compared to the earlier comparison on the CIFAR10 dataset, but it highlights MPFL's consistent performance. These results are visualized in Figure \ref{fig:10_units_with_noise}.

\begin{figure}
    \centering
    \includegraphics[width=1.0\linewidth]{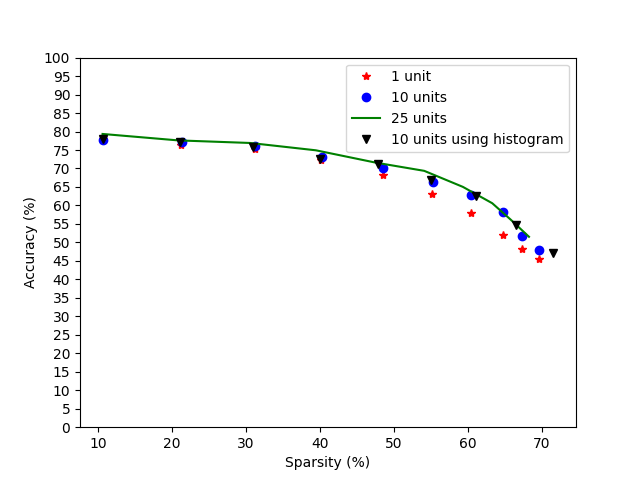}
    \caption{The accuracy versus sparsity levels at various node counts, as obtained in Experiment 2 while pruning the ResNet18 model using the ImageNet100 dataset through MPFL.}
    \label{fig:N_comparing}
\end{figure}

Finally, we conducted a comparison among varying numbers of nodes under noise and label mixing, incorporating our second method using histograms. This method entails inspecting the histogram and transferring high histogram percentages, requiring a minimum of $90\%$ agreement through the histogram. We observe that our two proposed methods of MPFL converge similarly while allowing different parameter flexibility in each. Additionally, we note that as the number of nodes increases, the stability of the voting algorithm also improves, attributed to the increased participation in the voting process. These results are depicted in Figure \ref{fig:N_comparing}. It is important to highlight that in this case, the improvement in relation to the number of nodes is not as significant, due to the robustness of the ResNet model against noise, but it is still present. 

\noindent
\textbf{Bandwidth-Efficiency:} In Table \ref{tab:sim2_bandwidth}, we evaluate the data transmission volumes among the algorithms. As a reminder, LTH follows a centralized approach, entailing a singular transfer where the entire communication-intensive, high-dimensional raw data is uploaded to the PS, followed by algorithm execution at the PS. We provide the total data transmission amount for this process. It is essential to note once again that LTH does not address privacy concerns, necessitating the non-sharing of direct data. It is presented solely as a performance benchmark. In contrast, MPFL and pruning-FL operate in a federated manner. They retain data locally at nodes while transmitting the updated model to the PS at each iteration. Here, we present the average transmitted data per iteration summed across all nodes, alongside the overall data transmitted throughout all iterations in the entire simulation runs.

Specifically, the centralized approach necessitates transmitting all raw data, involving the transmission of $130K$ images (1,300 images per class) and $5K$ images for the validation set (50 images per class), totaling $135K$ images with a shape of $(375,500,3)$, which is equivalent to $70.72$ GB. In the case of pruning-FL, each transmission deals with the model weights, encompassing a total of $129,176,036$ parameters for ResNet18. Assuming the use of float32, this leads to $492.76$ MB of transmitted data for each node per iteration. This calculation focuses solely on the bytes required to send weights and biases without additional information for each node. Across 10 iterations of pruning, we transmit a maximum of $4.81$ GB for each node. By not resending pruned weights and considering a total pruning rate not always at $10\%$ each iteration, the overall bandwidth used sums to $2.64$ GB for each node and $26.4$ GB in total.

In MPFL, every node transmits only a bit for each filter and a bit per value in the bias layer, totaling $26,000$ bits per iteration from each node. This translates to $3,250$ bytes for each iteration from each node, or $3.17$ KB per node. The overall bandwidth consumption required for the entire process is $32,500$ bytes for each node, amounting to less than $31.74$ KB for the entire process per node. By conserving bandwidth without retransmitting weight masks for layers that have already been pruned, we achieve $17,875$ bytes (or $17.45$ KB) for each node and $174.5$ KB for the entire network. As evident, for the larger dataset used in this experiment, pruning-FL proves to be more efficient compared to the centralized approach (LTH), as expected. MPFL demonstrates significant bandwidth savings over both algorithms, coupled with superior accuracy performance. 

These findings showcase the remarkable performance of MPFL, achieving substantial dimensionality reduction and resulting in significant bandwidth savings with superior accuracy compared to existing methods. Specifically, MPFL's bandwidth consumption is less than $1\%$ in small datasets (CIFAR10) and less than $10^{-3}\%$ in larger datasets (ImageNet100) compared to existing methods. These results position the MPFL algorithm as a highly promising solution for practical implementation in communication-constrained federated learning systems with limited hardware devices.

\begin{table}[htbp]
    \centering
    \caption{Total amount of transmitted data in
Experiment 2 for pruning the ResNet18 model using the ImageNet100 datase}
    \label{tab:sim2_bandwidth}
    \begin{tabularx}{\columnwidth}{Xcc}
        \toprule
        Algorithm & Single iteration (max) & Full simulation \\
        \midrule
        LTH & - & $70.72$ GB\\
        pruning-FL & $4.81$ GB & $26.4$ GB \\
        \textbf{Proposed MPFL} & \textbf{31.7 KB} & \textbf{174.5 KB} \\
        \bottomrule
    \end{tabularx}
\end{table}

\section{Conclusion}
\label{sec:conclusion}

We have developed a pioneer approach to FL tailored for scenarios involving edge devices with limited resources or constrained communication channel bandwidth. The algorithm, dubbed MPFL, seamlessly integrates pruning techniques with FL, effectively reducing the dimensionality of the transmitted model and optimizing resource usage.

In the experimental phase, our method exhibited similar scaling of pruning masks as other commonly used methods, yet with the remarkable advantage of consuming less than $1\%$ of the bandwidth required by these alternative approaches. Furthermore, our method demonstrated increased robustness in scenarios involving noisy nodes. Additionally, for high pruning sparsity, our model exhibited less damage to the accuracy of the network.

The results demonstrate the potential of MPFL to improve the efficiency of FL in resource-limited environments. By reducing the size of the transmitted model, MPFL successfully optimizes resource usage without compromising the quality of learning. Furthermore, MPFL maintains network stability through an iterative, consensus-based pruning process, thus enabling a robust FL system even in challenging scenarios.

\appendices{}

\bibliographystyle{ieeetr}
\bibliography{bibliography}

\end{document}